\documentclass[10pt,onecolumn,letterpaper]{article}
\usepackage{times}
\usepackage{graphicx}
\usepackage{amsmath}
\usepackage{amssymb}
\usepackage{multirow}

\newcommand{\Tr}{\mathrm{T}}

\newcommand{\RR}{{\mathbb{R}}}

\newcommand{\mm}[1]{\mathbf{#1}}

\newcommand{\dis}{\mathrm{dis}}

\begin{document}
%\doublespacing
\title{Affine-invariant geodesic geometry
         of deformable 3D shapes}

% for anonymous conference submission please enter N.N. or "Anonymous"
% instead of the author's name (and leave the affiliation blank) !!

%\author{Anonymous}

\author{Dan Raviv\\
Dept. of Computer Science\\
Technion, Israel\\
{\tt\small darav@cs.technion.ac.il}
\and
Alexander M. Bronstein\\
Dept. of Electrical Engineering\\
Tel Aviv University, Israel\\
{\tt\small bron@eng.tau.ac.il}
\and
Michael M. Bronstein\\
Dept. of Informatics\\
Universit{\`a} della Svizzera Italiana\\
Lugano, Switzerland\\
{\tt\small michael.bronstein@usi.ch}
\and
Ron Kimmel\\
Dept. of Computer Science\\
Technion, Israel\\
{\tt\small ron@cs.technion.ac.il}
\and
Nir Sochen\\
Dept. of Applied Mathematics\\
Tel Aviv University, Israel\\
{\tt\small sochen@math.tau.ac.il}
}

\maketitle

\begin{abstract}
Natural objects can be subject to various  transformations yet still preserve
 properties that we refer to as invariants.
Here, we use  definitions of affine invariant arclength for surfaces in $\RR^3$
 in order to extend the set of existing non-rigid shape analysis tools.
In fact, we show that by re-defining the surface metric as its equi-affine
version, the surface with its modified metric tensor can be treated as a canonical
 Euclidean object on which most classical Euclidean processing and analysis tools
 can be applied.
The new definition of a metric is used to extend the
 fast marching method technique for computing geodesic distances on surfaces,
 where now, the distances are defined with respect to an affine invariant arclength.
Applications of the proposed framework demonstrate its invariance, efficiency,
and accuracy in shape analysis.
\end{abstract}

%\maketitle

%%%%%%%%%%%%%%%%%%%%%%%%%%%%%%%%%%%%%%%%%%%%%
 \section{Introduction}
\label{sec:intro}

Modeling 3D shapes as Riemannian manifold is a ubiquitous approach in many shape analysis applications. %
In particular, in the recent decade, shape descriptors based on geodesic distances induced by a Riemannian metric have become popular. 
Notable examples of such methods are the canonical forms \cite{elad2003bis} and the Gromov-Hausdorff \cite{gro:GEOMETRY,mem:sap1:GEOMETRY,EffComIsoInv06} and the Gromov-Wasserstein \cite{dgh-spec-nordia} frameworks, used in shape comparison and correspondence problems. 
Such methods consider shapes as metric spaces endowed with a geodesic distance metric, and pose the problem of shape similarity as finding the minimum-distortion correspondence between the metrics. The advantage of the geodesic distances is their invariance to inelastic deformations (bendings) that preserve the Riemannian metric, which makes them especially appealing for non-rigid shape analysis. 
A particular setting of finding shape self-similarity can be used for intrinsic symmetry detection in non-rigid shapes \cite{raviv:bro:bro:kim:NRTL07,yang2008symmetry,lasowski-probabilistic,xu:symm}.

The flexibility in the definition of the Riemannian metric allows extending the invariance of the aforementioned shape analysis algorithms by constructing a geodesic metric that is also invariant to global transformations of the embedding space. 
A particularly general and important class of such transformations are the {\em affine} transformations, which play an important role in many applications in the analysis of images \cite{cordelia} and 3D shapes \cite{dinosaurs}.
Many frameworks have been suggested to cope with the action of the affine group in a global manner, 
trying to undo the affine transformation in large parts of a shape or a picture. While the theory of affine
invariance is known for many years \cite{Buchin1983AffineGeometry} and used for curves \cite{sapiro:affine:curve:94} and flows \cite{Sochen04}, no numerical constructions applicable to manifolds have been proposed.

In this paper, we construct an {\em (equi-)affine-invariant} Riemannian geometry for 3D shapes. 
By defining an affine-invariant Riemannian metric, we can in turn define affine-invariant geodesics, which result in a metric space with a stronger class of invariance.
This new metric allows us to develop efficient computational tools that
 handle non-rigid deformations as well as equi-affine transformations.
We demonstrate the usefulness of our construction in a range of shape analysis
 applications, such as shape processing, construction of shape descriptors, correspondence, and symmetry detection.

%Affine invariance is important in many applications in 
% the analysis of images \cite{cordelia} and 3D shapes \cite{dinosaurs}.
%%
%%Only five parameters are needed to construct the equi-affine transformation group in the plane \cite{kimmel2004:book},
%%and eleven in space.
%%, but revoking affinity depends on a good correspondence which is influenced by the transformation itself.
%%This ``chicken or the egg'' causality dilemma is the fundamental problem we encounter in affine geometry.
%%
%Many frameworks have been suggested to cope with the action of the affine group in a global manner, 
%trying to undo the affine transformation in large parts of a shape or a picture. While theory of affine
%invariance is known for many years \cite{Buchin1983AffineGeometry}, we encounter numerical schemes 
%from curves \cite{sapiro:affine:curve:94}, to equi-affine flows \cite{Sochen04} only recently.
%%Affinity is a local feature \cite{Buchin1983AffineGeometry,Sochen04}, and as such, an affine invariant flow
%%was suggested several years back.
%%
%%Here we provide a novel approach that combines local and global frameworks. 
%Here construct an affine-invariant Riemannian metric that allows us to redefine distances on any surface, with non-vanishing
% Principal curvature, and use it to calculate affine-invariant geodesics.
%This new geometry allows us to develop efficient computational tools that
% handle non-rigid deformations as well as equi-affine transformations.
%%
%We demonstrate the usefulness of our construction in a range of shape analysis
% applications.
\section{Background}
\label{sec:backgr}

We model a shape $(X,g)$ as a compact complete two-dimensional Riemannian manifold $X$ with a metric tensor $g$.  
The metric $g$ can be identified with an inner product $\langle \cdot, \cdot \rangle_x :T_x X \times T_x X \rightarrow \mathbb{R}$
on the tangent plane $T_x X$ at point $x$. 
We further assume that $X$ is embedded into $\RR^3$ by means of a regular map $\mm{x} : U \subseteq \RR^2 \rightarrow \RR^3$, 
so that the metric tensor can be expressed in coordinates as 
\begin{eqnarray}
 g_{ij} = \frac{\partial \mm{x}^\Tr}{\partial u_i}  \frac{\partial \mm{x} }{\partial u_j},
\end{eqnarray}
where $u_i$ are the coordinates of $U$.

The metric tensor relates infinitesimal displacements in the parametrization domain $U$ to displacement on the manifold
\begin{eqnarray}
 dp^2 = g_{11}{du_1}^2  + 2g_{12}du_1 du_2 + g_{22}{du_2}^2. 
\end{eqnarray}
This, in turn, provides a way to measure length structures on the manifold. Given a curve $C:[0,T] \rightarrow X$, its length can be expressed as 
\begin{eqnarray}
\ell(C) &=& \int_0^T \langle \dot{C}(t), \dot{C}(t) \rangle^{1/2}_{C(t)} dt, 
\end{eqnarray}
where $\dot{C}$ denotes the velocity vector.

\subsection{Geodesics}
%A special class of curves for which the velocity vector is parallel along the curve, are called {\em geodesics}. 
%Geodesic are extrema of the length functional $\ell(C)$. 
{\em Minimal geodesics} are the minimizers of $\ell(C)$, giving rise to the {\em geodesic distances}  
\begin{eqnarray}
\label{eq:geodesic}
d_X(x,x') = \min_{C \in \Gamma(x,x')} \ell(C)
\end{eqnarray}
where $\Gamma(x,x')$ is the set of all admissible paths between the points $x$ and $x'$ on the surface $X$, where due to completeness assumption, 
the minimizer always exists.

%One special set of curves derived from the metric are those with a local-length minimum, referred to as \emph{geodesics}.
%It can be shown that the velocity of a geodesic $x(p)$ is parallel along the curve. 

Structures expressible solely in terms of the metric tensor $g$ are called {\em intrinsic}. 
For example, the geodesic can be expressed in this way. The importance of intrinsic structures stems from the fact that they are invariant under isometric transformations 
(bendings) of the shape. In an isometrically bent shape, the geodesic distances are preserved -- a property allowing to use such structures as invariant shape descriptors \cite{elad2003bis}.

%Geometric elastic deformations which do not stretch or tear the shape, also known as bending, 
%are called \emph{intrinsic}. They are defined solely by the metric and do not depend on the ambient space.
%Hence, distances of curves on the surface, are influenced solely by the metric tensor, 
%and can be presented using the quadratic form
%\begin{eqnarray}
% dp^2 = g_{11}{du_1}^2  + 2g_{12}du_1 du_2 + g_{22}{du_2}^2.
%\end{eqnarray}

%One special set of curves derived from the metric are those with a local-length minimum, referred to as \emph{geodesics}.
%It can be shown that the velocity of a geodesic $x(p)$ is parallel along the curve, 
%
%which leads to the geodesics equation \cite{docarmo1976dgc} 
%\begin{eqnarray}
% \frac{ d^2 x^i} {dp^2} + \Gamma ^i_{jk} \frac {dx^j} {dp} \frac {dx^k}{dp} = 0,
%\end{eqnarray}
% where $\Gamma ^i_{jk}$ is the Christoffel symbol of the second kind.
%
%Not all geodesics represent the shortest path between two points on the surface, 
%but the shortest path is a geodesic curve.
%As a consequence, the shortest path is invariant to inelastic shape deformations, and
%is a good choice for describing non-rigid shapes capturing their properties that do not depend on the deformations.  

\subsection{Fast marching}
The geodesic distance $d_X(x_0,x)$ can be obtained as the viscosity solution to the {\em eikonal equation} $\| \nabla d\|_2 = 1$ 
%(i.e., the largest $d$ satisfying  $\| \nabla d\|_2 \leq 1$) 
with boundary condition at the source point $d(x_0)=0$. 
In the discrete setting, a family of algorithms for finding the viscosity solution of the discretized
eikonal equation by simulated wavefront propagation is called {\em fast marching
methods} \cite{Tsitsiklis95:FMM,KimmelSethian98:FMM}. 
On a discrete shape represented as a triangular mesh with $N$ vertices, the general structure of fast marching closely resembles that of the classical Dijkstra's algorithm for shortest path computation in graphs, with the main difference in the update step. Unlike the graph
case where shortest paths are restricted to pass through the graph edges, the
continuous approximation allows paths passing anywhere in the mesh triangles. For that reason, the value of $d(x_0,x)$ has to be computed from the
values of the distance map at two other vertices forming a triangle with $x$. 
Computation of the distance map from a single source point has the complexity of $\mathcal{O}(N\log N)$. 
%In
%order to guarantee consistency of the solution, all such triangles must have
%an acute angle at $x$. Obtuse triangles are split at a preprocessing stage by
%adding virtual connections to non-adjacent vertices.

%While there exist several methods to calculate shortest paths, in this paper we focus on a first order scheme called \emph{fast marching} (FMM)
%\cite{KimmelSethian98:FMM}.
%The main idea behind fast marching is to simulate a wavefront propagation on a triangular mesh, 
%associating the time of arrival of the front with the distance traveled, assuming a constant propagation speed. 

%{\bf [...] }

%
%This numerical procedure measures the distance from a point to the rest of the points on the surface. Computation of the geodesic distance between 
%a pair of points of a manifold sampled at $N$ points can be performed in linear $\OO(N)$ time 
%\cite{BroBroDviKimWebSIGGRAPH,Tsitsiklis95:FMM}. 
%

%In the proposed equi-affine scheme we considered the metric to be constant in each triangle. 
%We measured the length of each edge according to the equi-affine metric,
%and embedded the results into the FMM scheme. 

\section{Affine-invariant geometry}
\label{sec:affine}

An \emph{affine transformation} $\mm{x} \mapsto \mm{A}\mm{x} + \mm{b}$ of the three-dimensional Euclidean space
can be parametrized using twelve parameters: nine for the linear transformation $\mm{A}$, and additional three, $\mm{b}$,
for a translation, which we will omit in the following discussion.
Volume-preserving transformations, %(Figure \ref{Fig:affine_dogs}), 
known as \emph{special} or \emph{equi-affine} are restricted by $\det \mm{A} = 1$. 
Such transformations involve only eleven parameters.% (eight omitting translation).

%\begin{figure}[tpb]
%  \centering \includegraphics*[width=1\columnwidth]{images/affine_dogs.png}
%  \caption{ \label{Fig:affine_dogs} \small %An \emph{equi-affine transformation} $\mm{x} \mapsto \mm{A}\mm{x} + \mm{b}$ of a mesh in space ($\det \mm{A} = 1$). 
%  Example of an equi-affine transformation. 
%}
%\end{figure}

%
%In what follows, we are going to construct an equi-affine metric and sustitute it with the Euclidean one.

%substitute the Euclidean metric by its equi-affine invariant counterpart. 
%That, in turn, will induce an equi-affine-invariant Laplace-Beltrami operator and define equi-affine-invariant diffusion geometry.

The equi-affine metric can be defined through the parametrization of a curve on the surface
\cite{Axioms:Image:Proc:morel93,Proj:Inv:Bruck97,Buchin1983AffineGeometry,Best:Scale:Space:Caselles:96,Geometric:Evolutions:Olver:97,Sochen04}.
Let $C$ be a curve on $X$ parametrized by $p$.
By the chain rule,
\begin{eqnarray}
\frac{dC}{dp} &=& \mm{x}_1 \frac{d u_1}{d p} + \mm{x}_2 \frac{d u_2}{d p} \\
\frac{d^2C}{dp^2} &=& \mm{x}_1 \frac{d^2 u_1}{dp^2} + \mm{x}_2 \frac{d^2 u_2}{dp^2} + \mm{x}_{11} \left(\frac{du_1}{dp}\right)^2 +  2\mm{x}_{12}\frac{du_1}{dp} \frac{du_2}{dp}+\mm{x}_{22} \left(\frac{du_2}{dp}\right)^2, \nonumber
\label{eq:chain-rule}
\end{eqnarray}
where, for brevity, we denote $\mm{x}_i = \frac{\partial \mm{x}}{\partial u_i}$ and $\mm{x}_{ij} = \frac{\partial^2 \mm{x}}{\partial u_i \partial u_j}$.
As volumes are preserved under the equi-affine group of transformations,
 we define the invariant arclength $p$ through
\begin{eqnarray}
\label{eq:arclength}
 f(X) \det (\mm{x}_1,\mm{x}_2,C_{pp}) &=& 1,
\end{eqnarray}
where $f(X)$ is a normalization factor for parameterization invariance.
Plugging (\ref{eq:chain-rule}) into (\ref{eq:arclength}) yields
\begin{eqnarray}
\label{eq:arclength_1}
dp^2 &=&f(X)  \det (\mm{x}_1,\mm{x}_2, \mm{x}_{11} du_1^2 + 2\mm{x}_{12}du_1 du_2+\mm{x}_{22}du_2^2)\nonumber\\
&=&f(X) \left( \tilde g_{11} du_1^2 + 2 \tilde g_{12} du_1 u_2 + \tilde g_{22} du_2^2 \right).
\end{eqnarray}
where $\tilde g_{ij} =  \det (\mm{x}_1,\mm{x}_2, \mm{x}_{ij})$.

In order to evaluate $f(X)$ such that the quadratic form (\ref{eq:arclength_1}) will also be parameterization invariant, 
we introduce an arbitrary parameteriation $\bar {u}_1$ and $\bar {u}_2$, for which
 $\bar {\mm{x}}_i = \frac{\partial \mm{x}}{\partial \bar {u}_i}$ and $\bar {\mm{x}}_{ij} = \frac{\partial^2 \mm{x}}{\partial \bar {u_i} \partial \bar{u}_j}$.
The relation between the two sets of parameterizations can be expressed using the chain rule
\begin{eqnarray}
\label{formula:chain1}
 \bar X_1 = X_{\bar {u}_1} = X_{u_1} {u_1}_{\bar {u}_1} + X_{u_2} {u_2}_{\bar {u}_1} \\
 \bar X_2 = X_{\bar {u}_2} = X_{u_1} {u_1}_{\bar {u}_2} + X_{u_2} {u_2}_{\bar {u}_2}. \nonumber
\end{eqnarray}

It can shown \cite{Buchin1983AffineGeometry} using the Jacobian
\begin{eqnarray}
 J = \begin{pmatrix} {u_1}_{\bar {u}_1} & {u_2}_{\bar {u}_1} \\ {u_1}_{\bar {u}_2} & {u_2}_{\bar {u}_2} \end{pmatrix},
\end{eqnarray}
that
\begin{eqnarray}
 \label{formula:new_metric}
 \bar g_{11} d\bar u_1 ^2 + 2\bar g_{12} d\bar u_1 d\bar u_2 + \bar g_{22} d\bar u_2 ^2 = \left( \tilde g_{11} du^2 + 2 \tilde g_{12} dudv + \tilde g_{22}dv^2 \right)|J|,
\end{eqnarray}
and 
%\begin{eqnarray}
%\label{formula:new_metric2}
 $\bar g_{11} \bar {g}_{22} - \bar {g}_{12} ^2 = \left( \tilde g_{11} \tilde g_{22} - \tilde g_{12} ^2 \right) ^4$,
%\end{eqnarray}
where  $\bar{g}_{ij} = \det (\bar {\mm{x}}_1,\bar {\mm{x}}_2,\bar {\mm{x}}_{ij})$.
From (\ref{formula:new_metric}) % and (\ref{formula:new_metric2}) 
we evaluate the affine invariant parameter normalization
%\begin{eqnarray}
 $f(X) = \left|\bar {g}\right|^{-1/4}$, 
%\end{eqnarray}
and define an equi-affine pre-metric tensor \cite{Buchin1983AffineGeometry,Sochen04}
\begin{eqnarray}
 \label{eq:equi_metric}
  \hat{g}_{ij} &=& \bar {g}_{ij}  \left|\bar {g}\right|^{-1/4}.
\end{eqnarray}

The pre-metric tensor~(\ref{eq:equi_metric})  defines a true metric only for strictly
 convex surfaces \cite{Buchin1983AffineGeometry}; A simialr problem appeared in equi-affine curve evolution
where the flow direction was determined by the curvature vector. 
In two dimentions we can encounter non-positive definite metrics in concave, and hyperbolic points.
We propose fixing the metric by flipping the main axes of the operator, if needed.
In practice we restrict the eigenvalues of the tensor to be positive, and re-avaluate it.
From the eigendecomposition  $\hat{\mm{G}} = \mm{U} \mm{\Gamma}\mm{U}^\Tr$ of $\hat{g}$ in matrix notation, 
%\begin{eqnarray}
%\end{eqnarray}
where
$\mm{U}$ is orthonormal and $\mm{\Gamma} = \mathrm{diag}\{\gamma_1,\gamma_2 \}$, 
we compose a new  metric $\mm{G}$, such that 
%\begin{eqnarray}
 $\mm{G} = \mm{U} |\mm{\Gamma}| \mm{U}^\Tr$ 
%\end{eqnarray}
is positive definite and equi-affine invariant.

\begin{figure}[t!]
  \centering \includegraphics[width=\columnwidth]{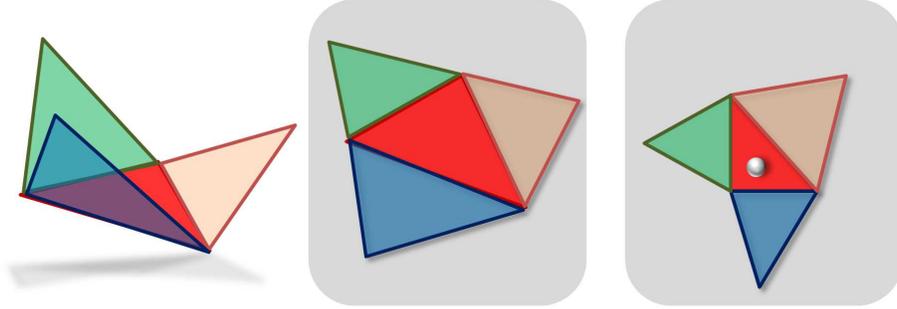}
   \caption{\label{fig:unfolding}
%\small{   %Left to right: part of a triangulated surface about a specific triangle.
   The three neighboring triangles together with the central one are unfolded flat to the plane.
   The central triangle is canonized into a right isosceles triangle while
   the rest of its three
   neighboring triangles follow the same planar affine transformation.
   Finally, the six surface coordinate values at the vertices are used to interpolate
   a quadratic surface patch from which the metric tensor is computed. %}
   }
\end{figure}
%
%Once all three coordinates functions are known, evaluating the equi-affine metric becomes trivial.

\begin{figure}[t!]

  \centering \includegraphics*[width=0.75\linewidth]{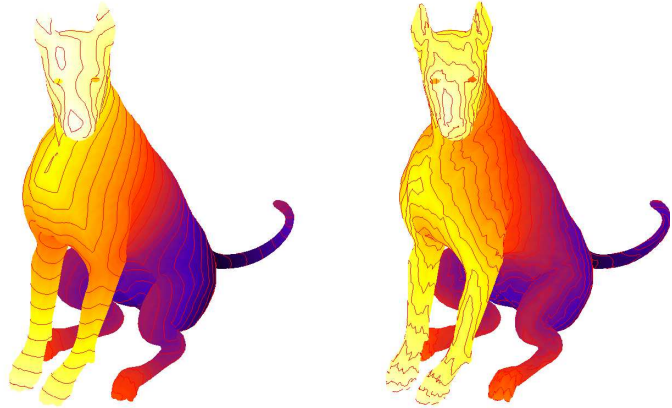}

 \caption{ \label{Fig:levelsets} Geodesic level sets of the distance function computed from the tip of the tail, 
using the standard (left) and the proposed equi-affine (right) geodesic metrics.}
\end{figure}

\begin{figure*}[th!]
\centering
  \begin{minipage}[b]{0.7\linewidth}
 %  \centering Affine transformation\\
   \centering \small \hspace{10mm}  Standard    \hspace{35mm} Equi-affine
   \centering \includegraphics*[width=1\linewidth]{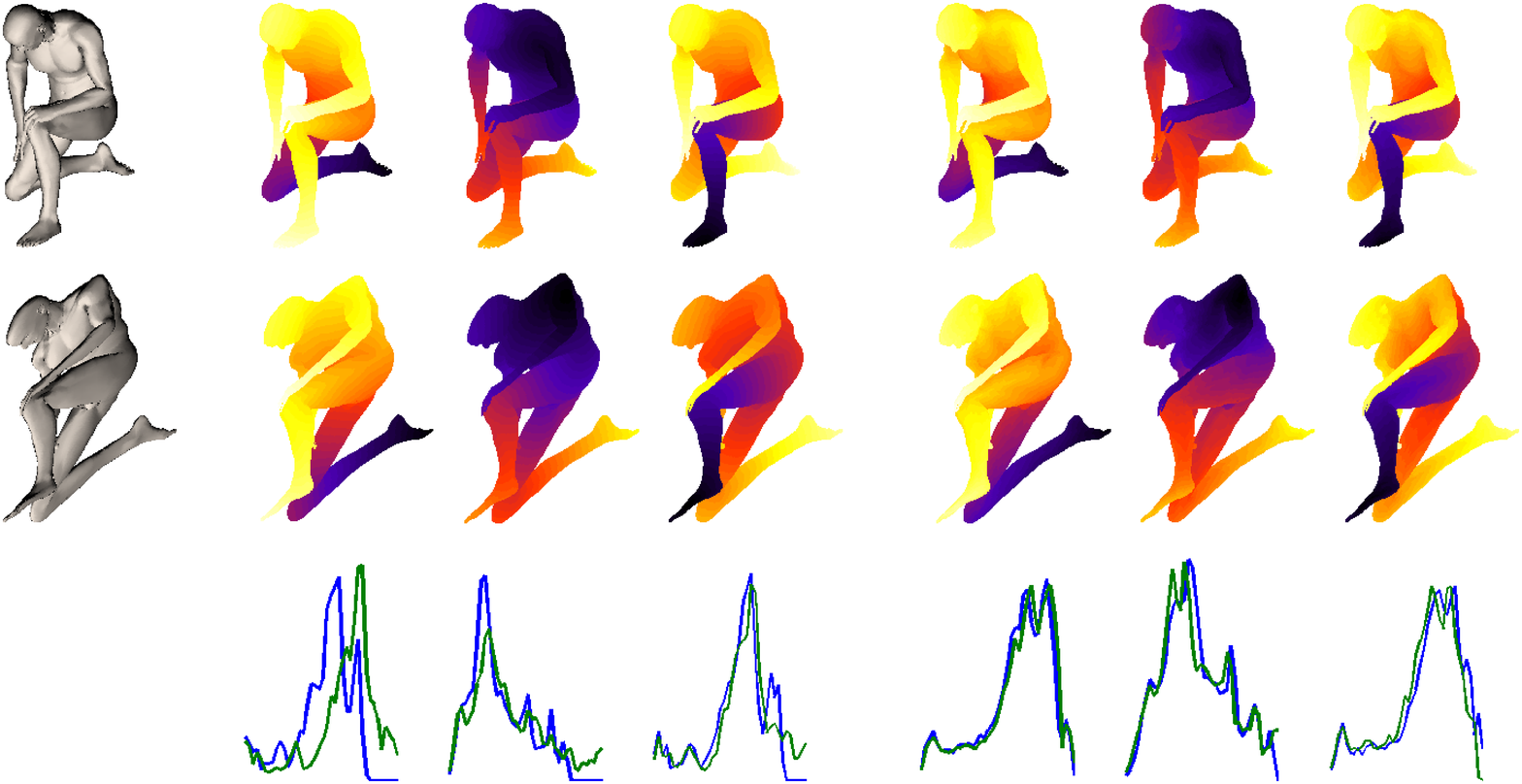}
   \vspace{5mm}
  \end{minipage}

  \begin{minipage}[b]{0.7\linewidth}
%  \centering  Isometric bending and Affine transformation\\
   \centering \small \hspace{10mm}  Standard    \hspace{35mm} Equi-affine
  \centering \includegraphics*[width=1\linewidth]{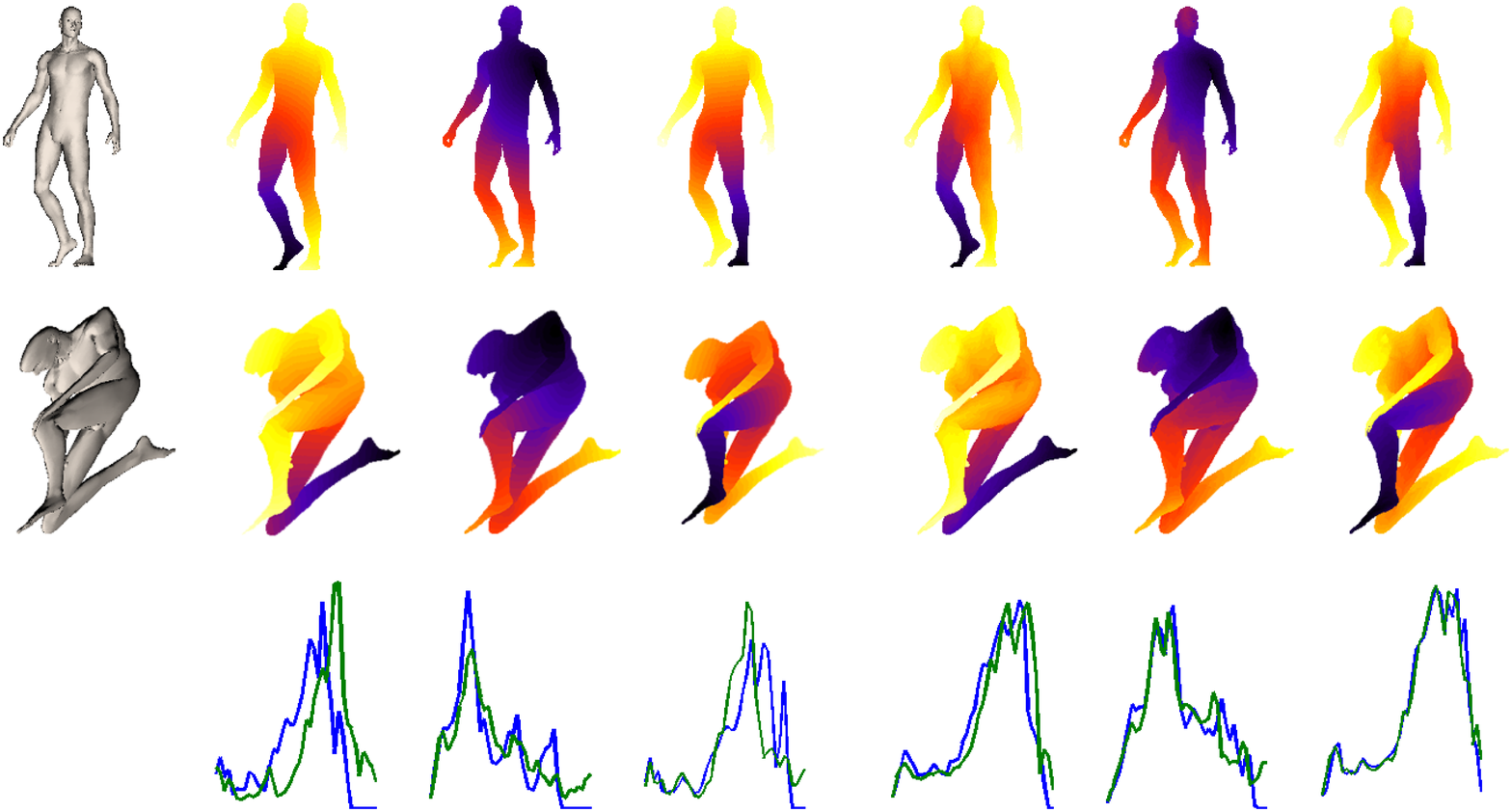}
  \end{minipage}

 \caption{ \label{Fig:geodesics} Distance maps from different source points calculated using the standard (second to fourth columns) and the proposed equi-affine geodesic metric (fifth to seventh columns) on a reference surface (first and third rows) and its affine (second row) and isometric deformation+affine transformation (fourth row). Thirds and sixth rows show the global histogram of geodesic distances before and after the transformation (green and blue curves).
 The overlap between the histograms is an evidence of invariance.
% While the equi-affine metric has to cope with this distortion alone, the Euclidean metric suffers from the combined error originated from the isometric
 %bending the the affine transformation.
 }

\end{figure*}

\section{Discretization}
\label{sec:num}

We model the surface $X$ as a triangular mesh, and construct three coordinate functions
$x(u,v)$, $y(u,v)$, and $z(u,v)$
for each triangle.
While this can be done practically in any representation, we use the fact that a triangle and
its three adjacent neighbors, can be unfolded to the plane, and produce a parameter domain.
The coordinates of this planar
 representation are used as the parametrization  with respect to which the first fundamental form coefficients are computed
at the barycenter of the simplex (Figure \ref{fig:unfolding}).
Using the six base functions $1$, $u$, $v$, $uv$, $u^2$ and $v^2$ we can
construct a second order polynom for each coordinate function.
This step is performed for every triangle of the mesh.

Calculating geodesic distances was well studied in past decades. Several fast and accurate numerical schemes
%\cite{KimmelSethian98:FMM,Sethian96:FMM,SpiraKimmel04:FMM,Yatziv06:FMM}
\cite{KimmelSethian98:FMM,sskgh2005fea,Yatziv06:FMM}
can be used off-the-shelf for this purpose.
We use FMM technique, after locally rescaling each edge according to the equi-affine metric.
%Since we assume the metric is constant in each triangle, we evaluate the edges' length by
% \begin{eqnarray}
%  {L_1}^2 &=& \begin{pmatrix}
%         1 & 0
%        \end{pmatrix}
% 	\begin{pmatrix}
%         g_{11} & g_{12} \\
% 	g_{21} & g_{22}
%        \end{pmatrix}
% 	\begin{pmatrix}
%         1 \\ 0
%        \end{pmatrix} = g_{11} \nonumber\\
%  {L_2}^2 &=& \begin{pmatrix}
%         0 & 1
%        \end{pmatrix}
% 	\begin{pmatrix}
%         g_{11} & g_{12} \\
% 	g_{21} & g_{22}
%        \end{pmatrix}
% 	\begin{pmatrix}
%         0 \\ 1
%        \end{pmatrix} = g_{22} \nonumber\\
%  {L_3}^2 &=& \begin{pmatrix}
%         1 & -1
%        \end{pmatrix}
% 	\begin{pmatrix}
%         g_{11} & g_{12} \\
% 	g_{21} & g_{22}
%        \end{pmatrix}
% 	\begin{pmatrix}
%         1 \\ -1
%        \end{pmatrix} \\
%  &=& g_{11} - 2g_{12} + g_{22}, \nonumber
% \end{eqnarray}
% \begin{eqnarray}
%  {L_1}^2 &=& 1\times g_{11} + 0\times g_{12} + 0\times g_{22} = g_{11} \nonumber\\
%  {L_2}^2 &=& 0\times g_{11} + 0\times g_{12} + 1\times g_{22} = g_{22} \nonumber\\
%  {L_3}^2 &=& 1\times g_{11}  -2\times g_{12} + 1\times g_{22} \nonumber\\
%   &=& g_{11} - 2g_{12} + g_{22}
% \end{eqnarray}

The (affine invariant) length of each edge is defined by $L^2(dx,dy) = g_{11}dx^2+2g_{12}dxdy+g_{22}dy^2.$
Specifically, for our canonical triangle with vertices at $(0,0),(1,0)$ and $(0,1)$ we have
\begin{eqnarray}
L^2_1 &=&  g_{11}(1-0)^2 + 2g_{12}(1-0)(0-0) + g_{22}(0-0)^2 \cr
 &=& g_{11} \cr
L^2_2 &=&  g_{11}(0-0)^2 + 2g_{12} (0-0)(1-0) +  g_{22}(1-0)^2 \cr
&=& g_{22} \cr
L^2_3 &=&  g_{11}(1-0)^2  +2 g_{12}(0-1)(1-0)  + g_{22}(1-0)^2\cr
  &=& g_{11} - 2g_{12} + g_{22}.
\end{eqnarray}  
Each edge may appear in more than one triangle. We found that the average length is a good approximation, assuming the triangle inequality holds.
In figures \ref{Fig:levelsets} and \ref{Fig:geodesics} we compare between geodesic distances induced by the standard and our affine-invariant metric.
\section{Results}
\label{sec:app}

\begin{figure*}[th!]
  \centering \includegraphics*[width=1\linewidth]{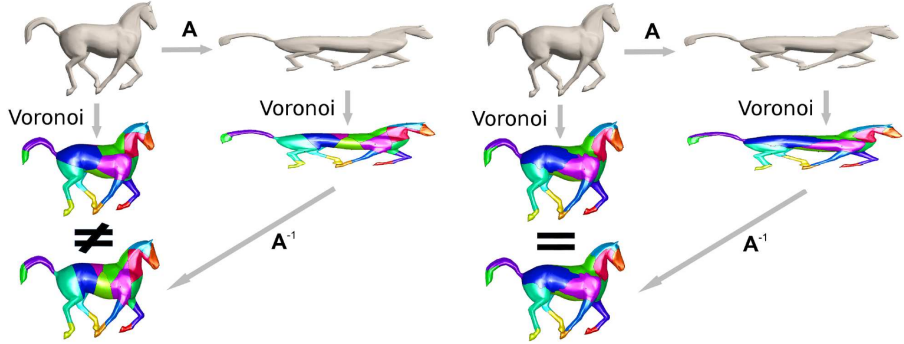}
  \caption{ \label{Fig:voronoi} Voronoi cells generated by a fixed set of $20$ points on a shape undergoing an equi-affine transformation.
The standard geodesic metric (left) and its equi-affine counterpart (right) were used. 
Note that in the latter case the tessellation commutes with the transformation.
}
\end{figure*}

The equi-affine metric can be used in many existing methods that process geodesic distances.
In what follows, we show several examples for embedding the new metric in known applications such as
voronoi tessellation, canonical forms, non-rigid matching and symmetry detection.

\subsection{Voronoi tessellation}
Voronoi tessellation is a partitioning of $(X,g)$ into disjoint open sets called Voronoi cells.
%decomposition of a metric space $(X,g)$ into non-overlapping parts $V_i$ such that the $i-$th cell contains all points
A set of $k$ points $\left(x_i \in X\right)_{i=1}^k $ on the surface define the Voronoi cells $\left(V_i\right)_{i=1}^k$
such that the $i$-th cell contains all points on $X$ closer to $x_i$ than to any other $x_j$ in the sense of the metric $g$.
%(in other words, a point $x\in X$ belongs the $i$'th cell if $d_X(x,x_i) \le d_X(x,x_j)\,\, j\neq i$. where $d_X : X \times X\rightarrow \mathbb{R}$ is the distance metric.
%
%map induced by the Riemannian metric tensor $g$. Boundary points which have a similar distance to more than one centroid are considered to be in none or all of the cells.
%
Voronoi tessellations created with the equi-affine metric commute with equi-affine transformations as visualized in Figure~\ref{Fig:voronoi}.

\subsection{Canonical forms}
%The similarity of two non-rigid
Methods considering shapes as metric spaces with some intrinsic (e.g. geodesic) distance metric is an important class of approaches in shape analysis. Geodesic distances are particularly appealing due to their invariance to inelastic deformations that preserve the Riemannian metric.

Elad and Kimmel \cite{elad2003bis} proposed a shape recognition algorithm based on embedding the metric structure of a shape $(X,d_X)$ into a 
low-dimensional Euclidean spaces. Such a representation, referred to as {\em canonical form}, 
reduces the number of degress of freedom by tanslating all deformations into a much simple Euclidean isometry group. 
%for example, the Hausdorff distance can be used to compare two canonical forms.

Given a shape sampled at $N$ points and an $N\times N$ matrix of pairwise geodesic distances, the computation of the canonical form consists of finding a configuration of $N$ points $z_1,\hdots,z_N$ in $\RR^m$ such that $\|z_i - z_j\|_2 \approx d_X(x_i,x_j)$.
This problem is known as {\em multidimensional scaling} (MDS) and can be posed as a non-convex least-squares optimization problem of the form
\begin{eqnarray}
\{z_1,\hdots,z_N\} &=& \mathop{\mathrm{argmin}}_{z_1,\hdots,z_N} \sum_{i>j} | \|z_i - z_j\|_2 - d_X(x_i,x_j) |^2.
\end{eqnarray}

\begin{figure*}[tpb]
\centering
% \begin{minipage}[b]{0.55\linewidth}
 %  \centering Affine transformation\\
  % \centering \small \hspace{10mm}  Affine transformation strength
   \includegraphics[width=0.35\linewidth]{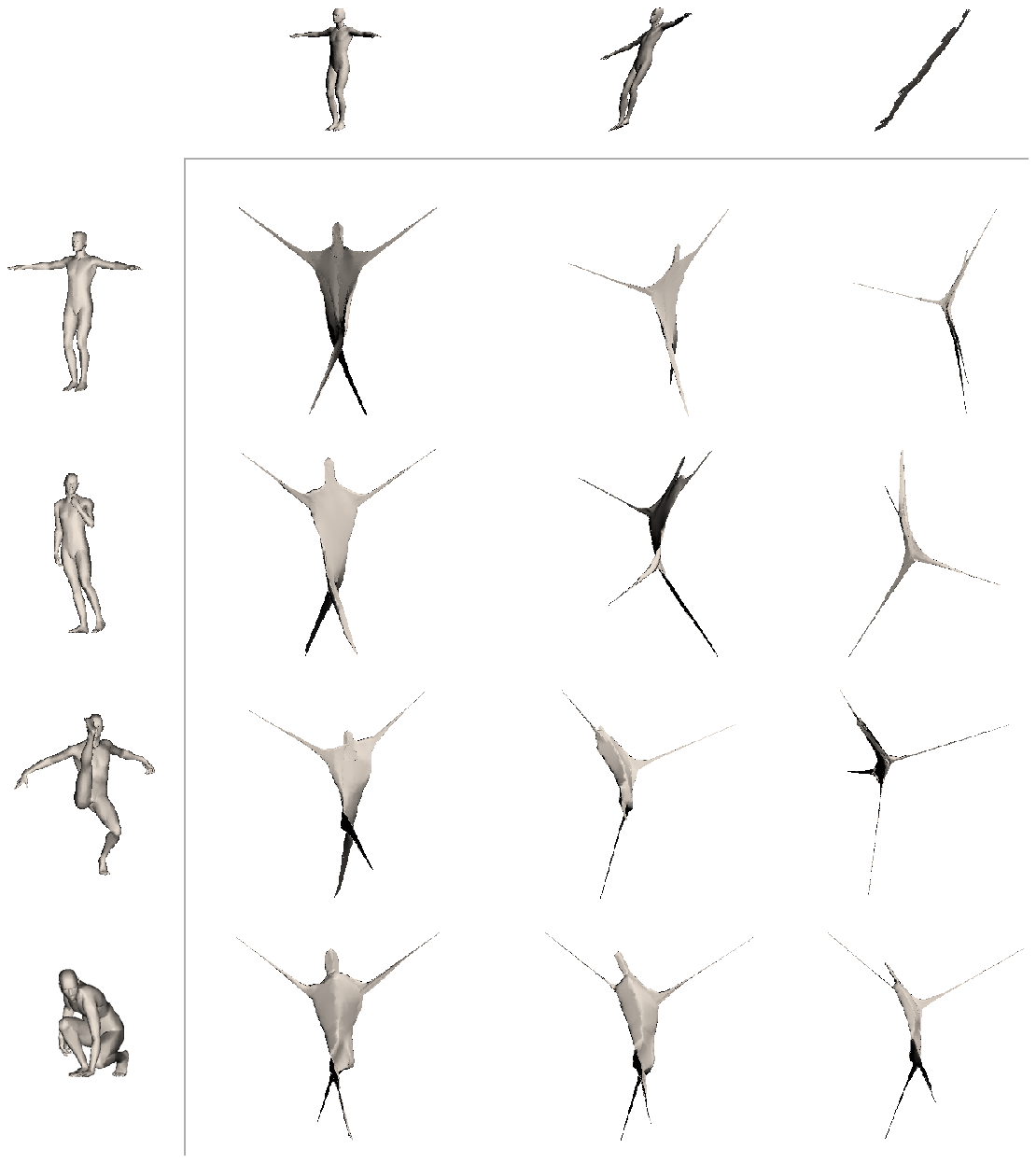}
  %\end{minipage}
%
\hspace{15mm}
%
%  \begin{minipage}[b]{0.4\linewidth}
%  \centering  Isometric bending and Affine transformation\\
   %\centering \small \hspace{20mm}  Affine transformation strength
  \includegraphics[width=0.278\linewidth]{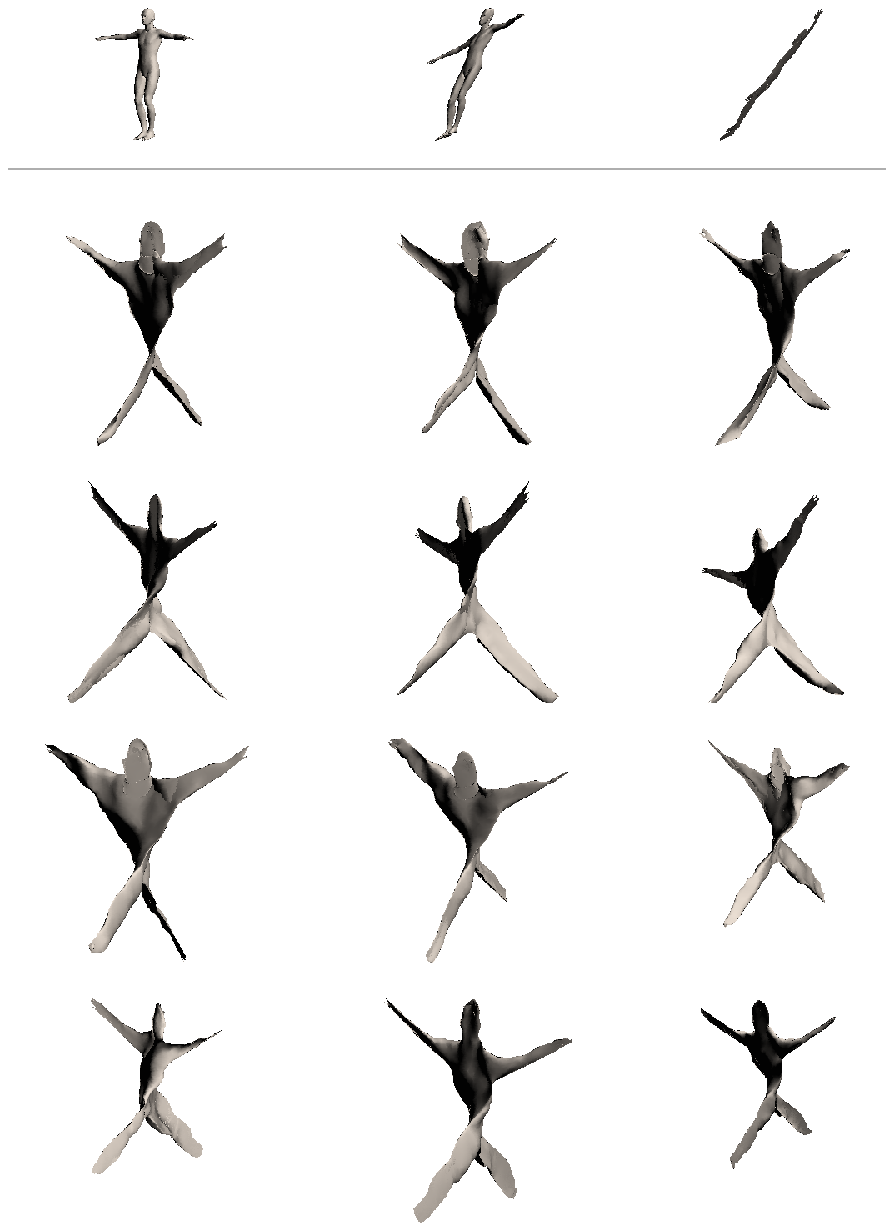}
%  \end{minipage}

 \caption{ \label{Fig:canonical_form} Embedding into $\mathbb{R}^3$ of a human shape and its equi-affine transformations of varying strength.
 Classical scaling was used with a matrix of geodesic (left) and equi-affine geodesic (right) distances. In the latter case, canonical forms remain
 approximately invariant up to a rigid transformation. }

\end{figure*}

The invariance of the canonical form to shape transformations depends on the choice of the distance metric $d_X$.
Figure~\ref{Fig:canonical_form} shows an example of a canonical form of the human shape undergoing different bendings and affine transformations of varying strength. The canonical form was computed using the geodesic and the proposed equi-affine distance metric.
One can clearly see the nearly perfect invariance of the latter. Such a strong invariance allows to compute correspondence of full shapes 
under a combination of inelastic bendings and affine transformations.
 %equi-affine canonical form
%remains invariant to the the affine transformation upto rigid symmetries, while using the traditional metric the canonical form is not unique.

%\begin{figure}[tpb]
%
%  \centering \includegraphics*[width=.9\linewidth]{images/canonical_forms.png}
%
% \caption{ \label{Fig:canonical_form} \small Embedding in $\mathbb{R}^3$ of a human body in different poses subject to different affine transformations. The equi-affine canonical form
%remains invariant to the the affine transformation upto rigid symmetries, while using the traditional metric the canonical form is not unique.
%}
%
%\end{figure}

\subsection{Non rigid matching}

Two non-rigid shapes $X, Y$ can be considered similar if there exists an isometric {\em correspondence} $\mathcal{C} \subset X\times Y$ between them, such that $\forall x \in X$ there exists $y \in Y$ with $(x,y) \in \mathcal{C}$ and vice-versa, and $d_X(x,x') = d_Y(y,y')$ for all $(x,y), (x',y') \in \mathcal{C}$, where $d_X, d_Y$ are geodesic distance metrics on $X, Y$.
In practice, no shapes are truly isometric, and such a correspondence rarely exists; however, one can attempt finding a correspondence minimizing the metric {\em distortion},
\begin{eqnarray}
\dis(\mathcal{C}) = \mathop{\max_{(x,y) \in \mathcal{C}}}_{(x',y') \in \mathcal{C}} |d_X(x,x') - d_Y(y,y')|.
\end{eqnarray}
The smallest achievable value of the distortion is called the {\em Gromov-Hausdorff distance} \cite{bur:bur:iva:01:GEOMETRY} between the metric spaces $(X,d_X)$ and $(Y,d_Y)$,
\begin{eqnarray}
\label{gromov_hausdorf}
d_\mathrm{GH}(X,Y) = \frac{1}{2} \inf_\mathcal{C} \dis(\mathcal{C}),
\end{eqnarray}
and can be used as a criterion of shape similarity.

The choice of the distance metrics $d_X, d_Y$ defines the invariance class of this similarity criterion.
Using geodesic distances, the similarity is invariant to inelastic deformations.
Here, we use geodesic distances induced by our equi-affine Riemannian metric tensor, which gives additional invariance to affine transformations of the shape.

\begin{figure}[tpb]
  \centering \includegraphics*[width=.9\linewidth]{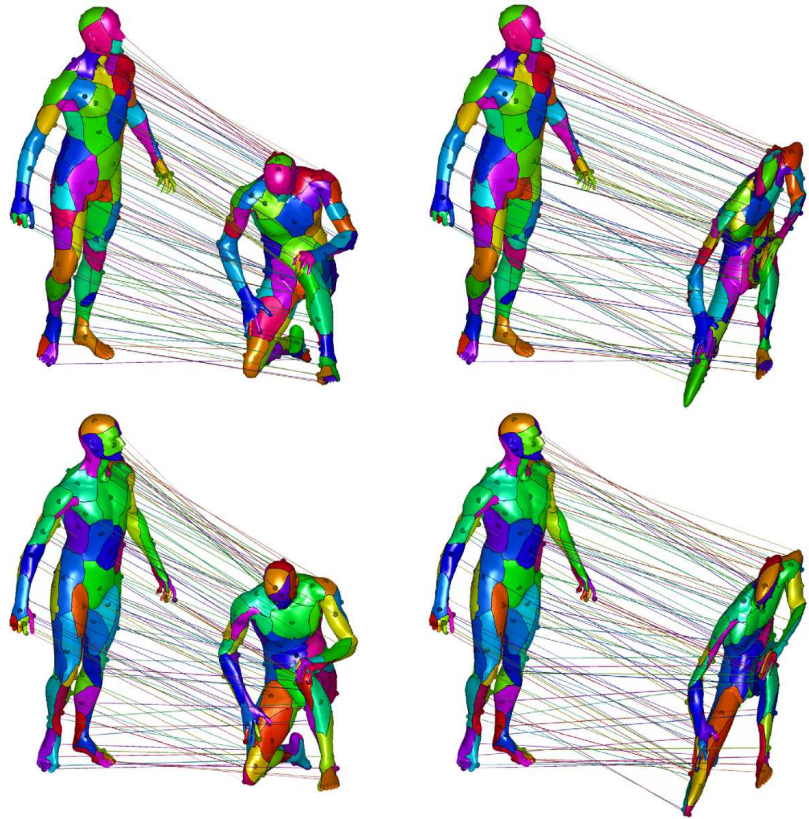}
  \caption{ \label{Fig:gmds-matching} The GMDS framework is used to calculate correspondences between a shape and its isometry (left) and
  isometry followed by an equi-affine transformation (right).
  Matches between shapes are depicted as identically colored Voronoi cells.
Standard distance (first row) and its equi-affine-invariant counterpart (second row) are used as the metric structure
in the GMDS algorithm. Inaccuracies obtained in the first case are especially visible in the legs and arms.
}
\end{figure}

Bronstein \emph{et al.} \cite{EffComIsoInv06} showed how (\ref{gromov_hausdorf}) can be efficiently approximated using an optimization algorithm in the spirit of multidimensional scaling (MDS), referred to as generalized MDS (GMDS).
%framework  in
%a framework referred to as Multi-Dimensional Scaling (GMDS).
Since the input of this numeric framework are geodesic distances between mesh points, all is needed to obtain an equi-affine GMDS is one additional step where we substitute the geodesic distances with their equi-affine equivalents.
Figure~\ref{Fig:gmds-matching} shows the correspondences obtained between an equi-affine transformation of a shape using the standard and the equi-affine-invariant versions of the geodesic metric.

\subsection{Intrinsic symmetry}

Raviv \emph{et al.}  \cite{raviv:bro:bro:kim:NRTL07}  introduced the notion of {\em intrinsic symmetries} for non-rigid shapes as self-isometries of a shape with respect to a deformation-invariant (e.g. geodesic) distance metric.
These self-isometries can be detected by trying to identify local minimizers of the metric distortion or other methods proposed in follow-up publications \cite{Ovs:Sun:Guibas:GloSym:08,yang2008symmetry,lasowski-probabilistic,xu:symm}.

\begin{figure*}[tpb]
  \centering \includegraphics*[width=1\linewidth]{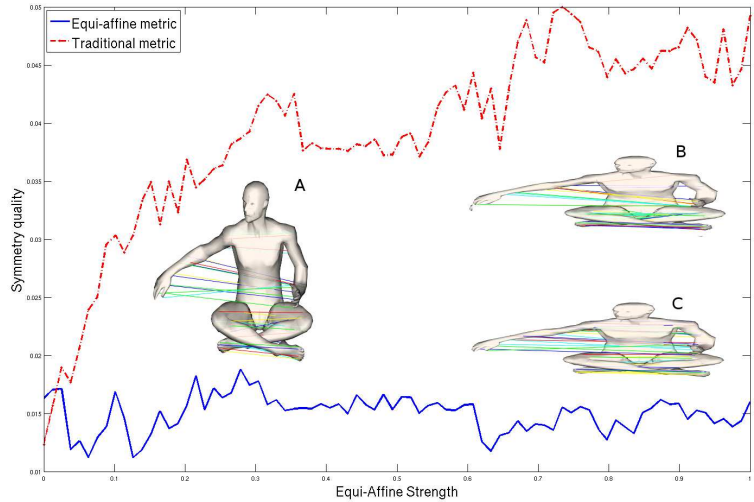}
  \caption{ \label{Fig:symmetry-matching} As the affine transformation becomes stronger, the quality
of the symmetry detection decreases (B) when the standard geodesic metric is used. On the other hand, detection quality
 is hardly affected (C) by the transformations when using the equi-affine geodesic metric.
}
\end{figure*}

Here, we adopt the %\cite{FullPartialSym:Raviv:BBK:2009}
framework of \cite{raviv:bro:bro:kim:NRTL07} for equi-affine intrinsic symmetry detection.
Such symmetries play an important role in paleontological applications \cite{dinosaurs}.
Equi-affine intrinsic symmetries are detected as local minima of the distortion, where the equi-affine geodesic distance metric is used.
Figure~\ref{Fig:symmetry-matching} shows that using the traditional metric we face a decrease in accuracy
of symmetry detection as the affine transformation becomes stronger 
(the accuracy is the average geodesic mismatch with relation to the ground-truth symmetry).
Such a decrease does not occur using the equi-affine metric.

\section{Conclusions}
\label{sec:concl}

% We introduced equi-affine-invariant geodesic distances,
%  and showed that it can be utilized to construct affine-invariant geodesics.
% %
% Performance of the proposed tools was demonstrated on different applications.
% %
%  Currently we are using a first order scheme to evaluate the affine metric. In the future we plan to improve 
% the numeric framework to achieve better accuracy.
% 
We introduced a numerical machinery for computing 
 equi-affine-invariant geodesic distances.
%, and showed that it can be utilized to construct affine-invariant
%geodesics. % We do not show this here! 
%Performance of 
The proposed tools were applied to % was demonstrated on different 
 applications, like symmetry detection, finding correspondence, and
canonization for efficient 
 shape matching. 
We have extended the ability to analyze approximately isometric objects, 
 like articulated objects, by treating affine deformations as well as
non-rigid ones.
As our analysis is based on local geometric structures,
 the affine group could in fact act locally and vary smoothly in space as
long as it is 
 encapsulated within our approximation framework.
We intend to explore this direction and further enrich the set of
transformations 
 one can handle within the scope of metric geometry in the future.

\section{Acknowledgements}
This research was supported in part by The Israel Science Foundation (ISF) grant number 623/08, 
and by the USA Office of Naval Research (ONR) grant.

% 
% \begin{figure}[h!]
% \includegraphics[width=\columnwidth]{template_figure.pdf}
% \caption{Example figure.} \label{fig:example_figure}ñ
% \end{figure}

%-------------------------------------------------------------------------
%\section{References}
\bibliographystyle{ieee}
\bibliography{bib_affine,bib_art,bib_extra}

\end{document}